\documentclass[letterpaper, 10 pt, journal, twoside]{IEEEtran}
\usepackage{graphicx}
\usepackage{mathptmx} 
\usepackage{times} 
\usepackage{amsmath} 
\usepackage{amssymb}  
\usepackage{diagbox}
\usepackage{array}
\usepackage{booktabs}
\usepackage{bm}
\newcommand{\specialcell}[2][c]{%
  \begin{tabular}[#1]{@{}c@{}}#2\end{tabular}}
\newcommand{\R}{\mathbf{R}}
\usepackage{xcolor}
\newcommand{\yufei}[1]{{#1}}
  \newcommand{\etal}{\textit{et al}.}
  \usepackage{cite}

\usepackage{microtype}
\renewcommand{\baselinestretch}{0.95}

\ifCLASSINFOpdf
\else
\fi
\begin{document}
%
\title{Visual Haptic Reasoning: Estimating Contact Forces by Observing Deformable Object Interactions}
%
%
%

\author{Yufei Wang, David Held, and Zackory Erickson%
\thanks{Manuscript received: February, 25, 2022; Revised June, 1, 2022; Accepted July, 28, 2022.}
\thanks{This paper was recommended for publication by Editor Angelika Peer upon evaluation of the Associate Editor and Reviewers' comments.
This work was supported by the National Science Foundation under Grant No. IIS-2046491, and LG Electronics. Corresponding author: Yufei Wang.} 
\thanks{The authors are with the Robotics Institute, Carnegie Mellon University, Pittsburgh, PA, USA. Emails: yufeiw2@andrew.cmu.edu, dheld@andrew.cmu.edu, zackory@cmu.edu}
\thanks{Digital Object Identifier (DOI): see top of this page.}
}
%
%

\markboth{IEEE Robotics and Automation Letters. Preprint Version. Accepted July, 2022}
{Wang \MakeLowercase{\textit{et al.}}: Visual Haptic Reasoning: Estimating Contact Forces by Observing Deformable Object Interactions} 

%



\maketitle

\begin{abstract}
Robotic manipulation of highly deformable cloth presents a promising opportunity to assist people with several daily tasks, such as washing dishes; folding laundry; or dressing, bathing, and hygiene assistance for individuals with severe motor impairments.
In this work, we introduce a formulation that enables a collaborative robot to perform visual haptic reasoning with cloth---the act of inferring the location and magnitude of applied forces during physical interaction.
We present two distinct model representations, trained in physics simulation, that enable haptic reasoning using only visual and robot kinematic observations.
We conducted quantitative evaluations of these models in simulation for robot-assisted dressing, bathing, and dish washing tasks, and demonstrate that the trained models can generalize across different tasks with varying interactions, human body sizes, and object shapes. 
We also present results with a real-world mobile manipulator, which used our simulation-trained models to estimate applied contact forces while performing physically assistive tasks with cloth. Videos can be found at our project webpage.\footnote{https://sites.google.com/view/visualhapticreasoning/home}

\end{abstract}

\begin{IEEEkeywords}
Physically Assistive Devices; Deep Learning for Visual Perception; Perception for Grasping and Manipulation
\end{IEEEkeywords}

%
\IEEEpeerreviewmaketitle

\section{INTRODUCTION}

\IEEEPARstart{R}{obotic} manipulation of highly deformable cloth presents a promising opportunity to assist people with many tasks, such as assisting an older adult with muscle atrophy or a physical disability to get dressed~\cite{erickson2018deep}, bathing and hygiene assistance with a washcloth or towel~\cite{erickson2020assistive}, cleaning dishes with a dish towel, folding laundry~\cite{weng2022fabricflownet, hoque2021visuospatial}, or bed making~\cite{seita2018deep, puthuveetil2022bodies}.  In each of these scenarios, it can be helpful for a robot to infer how cloth interacts with and applies forces to objects it makes contact with. For example, the applied force between a gown and human body can inform a robot if the gown is getting caught during assistive dressing, and the force between a washcloth and the cleaning surface can tell if the dirt on the surface is successfully removed during assistive bathing or dish cleaning. Besides inferring such task execution status, knowing the applied force is also vital to prevent harm and discomfort during physical interactions between robots and humans in these tasks.

\begin{figure}[t]
    \centering
    \includegraphics[width=.9\columnwidth]{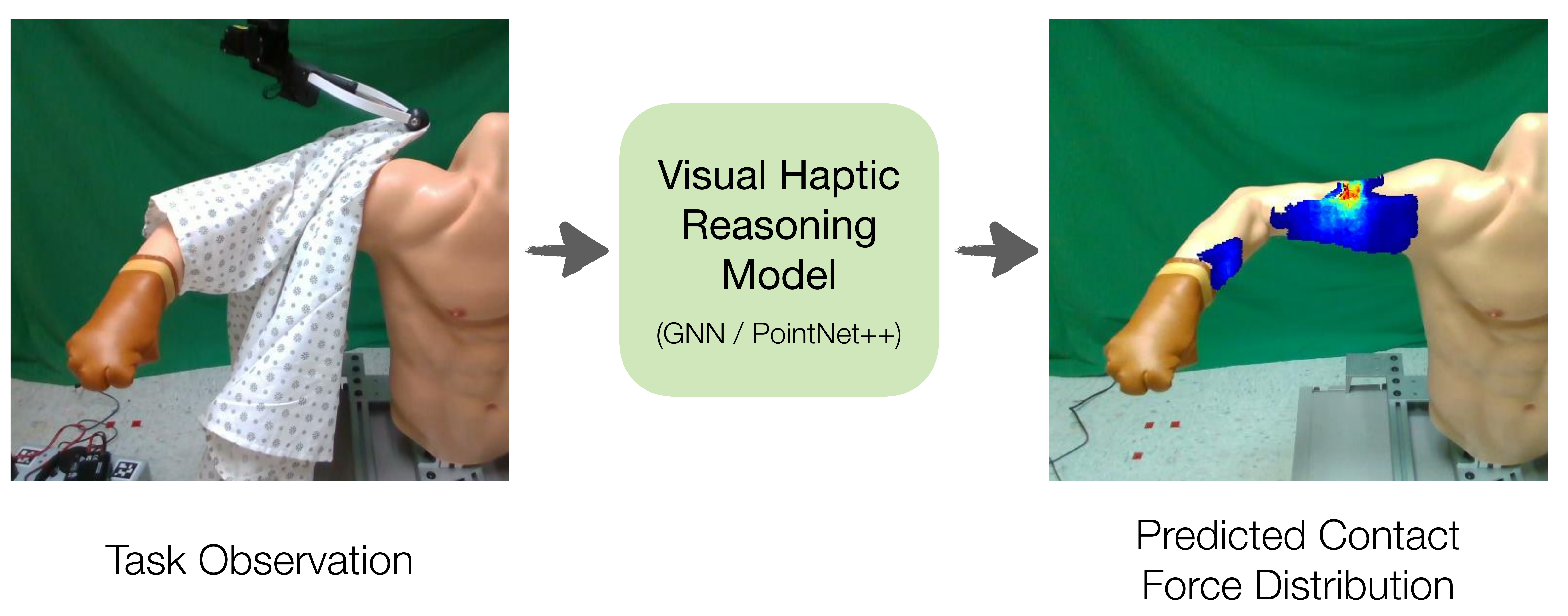}
    \vspace{-0.2cm}
     \caption{An illustration of the predicted contact force distributions by the proposed visual haptic reasoning model as a deformable gown physically interacts with a manikin in the robot-assisted dressing task. Note that the model is trained entirely in physics simulation. 
     }
    \vspace{-0.5cm}
    \label{fig:draw}
\end{figure}

In this paper, we introduce methods that enable collaborative robots to perform \emph{haptic reasoning} with cloth by using only \yufei{point cloud observations} and robot kinematics. As shown in Fig.~\ref{fig:draw}, \emph{haptic reasoning} consists of inferring the distribution of applied forces as cloth physically interacts with other objects. 
Prior work has introduced methods for estimating contact forces purely from visual feedback as rigid objects undergo contact with robot end-effectors or human hands at a few discrete points~\cite{pham2015towards, hwang2017inferring, ehsani2020use}. In contrast, cloth lacks a succinct state representation, has inherently high-dimensional non-linear dynamics, and has contacts with objects over large surface areas, which collectively presents unique challenges for estimating the location and magnitude of applied forces as cloth interacts with other objects in the environment.

To overcome these challenges, we take a data-driven approach with physics simulation to model the physical interactions between cloth and other objects. We present two model formulations for doing haptic reasoning during cloth manipulation.
The first introduces a graph neural network (GNN)~\cite{battaglia2018relational} architecture that encodes the local interactions between the cloth and object in contact, whereas the second approach uses a PointNet++~\cite{qi2017pointnet++} architecture. Both models form a mapping from a point cloud observation of the task to a 3D representation of the applied forces on an object.
We conduct quantitative evaluations in physics simulation and contrast the predicted force distributions of both model formulations during robot-assisted dressing, bathing, and dish washing tasks.
We perform ablation studies on these models and evaluate generalization performance as cloth interacts with a wide distribution of human body sizes and object shapes.
Finally, through cloth manipulation studies in the real world, we demonstrate that haptic reasoning models trained in physics simulation can be transferred to a real-world mobile manipulator to infer applied contact forces.

In summary, we make the following contributions:
\begin{itemize}
    \item We introduce model formulations for visual haptic reasoning, which enable a robot with visual sensing to infer the force distributions that cloth applies onto other objects during manipulation.
    \item We performed analysis of these haptic reasoning models in physics simulation across a number of tasks including robot-assisted dressing, bathing, and dish washing.
    \item We evaluate haptic reasoning in the real world and demonstrate these models with a mobile manipulator performing physically assistive tasks.
\end{itemize}

\section{RELATED WORK}

\subsection{Deformable cloth manipulation for robotic assistance}

Several assistive robotic tasks involve manipulating deformable objects like cloth around the human body. Examples include dressing assistance with hospital gowns, jackets, scarfs~\cite{erickson2020assistive, erickson2018deep, zhang2019probabilistic, erickson2021characterizing, gao2016iterative}; bathing assistance with a towel~\cite{erickson2020assistive, king2010towards},
picking and placing garments on hangers~\cite{antonova2021dynamic, zhang2020learning}, laundry folding~\cite{cusumano2011bringing, maitin2010cloth,weng2022fabricflownet, xu2022dextairity, ha2022flingbot, hoque2021visuospatial, lin2022learning, triantafyllou2011vision, osawa2007unfolding, seita2020deep}, and bedding assistance~\cite{puthuveetil2022bodies, seita2018deep}.
Prior work~\cite{yu2017haptic} has showed how a robot could predict if an end effector trajectory would succeed in dressing a hospital gown sleeve onto a person by leveraging force measurements at the end effector. 
In contrast to these prior works, we introduce a methodology for an assistive robot to infer the \yufei{location and magnitude of the forces} that cloth applies onto the human body using \yufei{point cloud observations}.

\subsection{Estimating Contact from Vision}
Several previous works~\cite{ehsani2020use, hwang2017inferring, pham2015towards, magrini2014estimation, zhu2016inferring, erickson2017does, brahmbhatt2019contactdb, haouchine2018vision, mendizabal2019force} have explored estimating  contact points and forces between objects in contact purely from vision using RGB, RGB-D or thermal images. Most of them focus on interactions between human hands and non-deformable objects~\cite{ehsani2020use, pham2015towards, brahmbhatt2019contactdb}, or between a rigid robot end effector and a non-deformable object~\cite{hwang2017inferring, magrini2014estimation, liang2021contact}, \yufei{or between a rigid robot tool and a deformable organ~\cite{haouchine2018vision, mendizabal2019force}}, where there are only a few points of contact.
In contrast to these prior works, our work focuses specifically on estimating contact forces when manipulating deformable cloth around other rigid objects, 
which results in hundreds of points of contact and applied forces across a human body or object surface.
Zhu~\etal~\cite{zhu2016inferring} are able to predict contact forces on every human mesh vertex when the human sits on a chair from RGB-D images, by using Finite Element Method (FEM) and modeling the human as a soft body. 
This approach is intractable for a collaborative robot that must make predictions in real time when physically interacting with people.
We instead use neural network models trained entirely in physics simulation to predict the force that cloth applies onto other objects, which is much faster as it only needs a single forward pass of the network.

\subsection{Estimating Contact Force during Interaction Involving Deformable Objects}
One prior work~\cite{erickson2017does} most relevant  to ours estimates the contact forces between a hospital gown and human limbs during assistive dressing. 
The trained models rely on manual discretization of the human limb into a fixed set of contact locations, which limits generalization. 
In a following work~\cite{erickson2018deep} the estimated forces are used in model predictive control for assistive dressing. 
Instead, by using \yufei{point cloud observations}, we need no such discretization for the object and our trained model generalizes across different assistive tasks, human shapes and object sizes.
Clever~\etal~\cite{clever2020bodies} used physics simulation to compute the pressure of a human lying on a deformable bed, and trained a neural network model for estimating human pose in bed.
We also use physics simulation to get high-resolution force distributions, but we focus on estimating the forces applied by deformable cloth to other objects in tasks such as robot-assisted dressing and bathing, instead of the forces applied from a lying rigid human to a deformable bed.

\begin{figure*}[t]
    \centering
    \includegraphics[width=.6\textwidth]{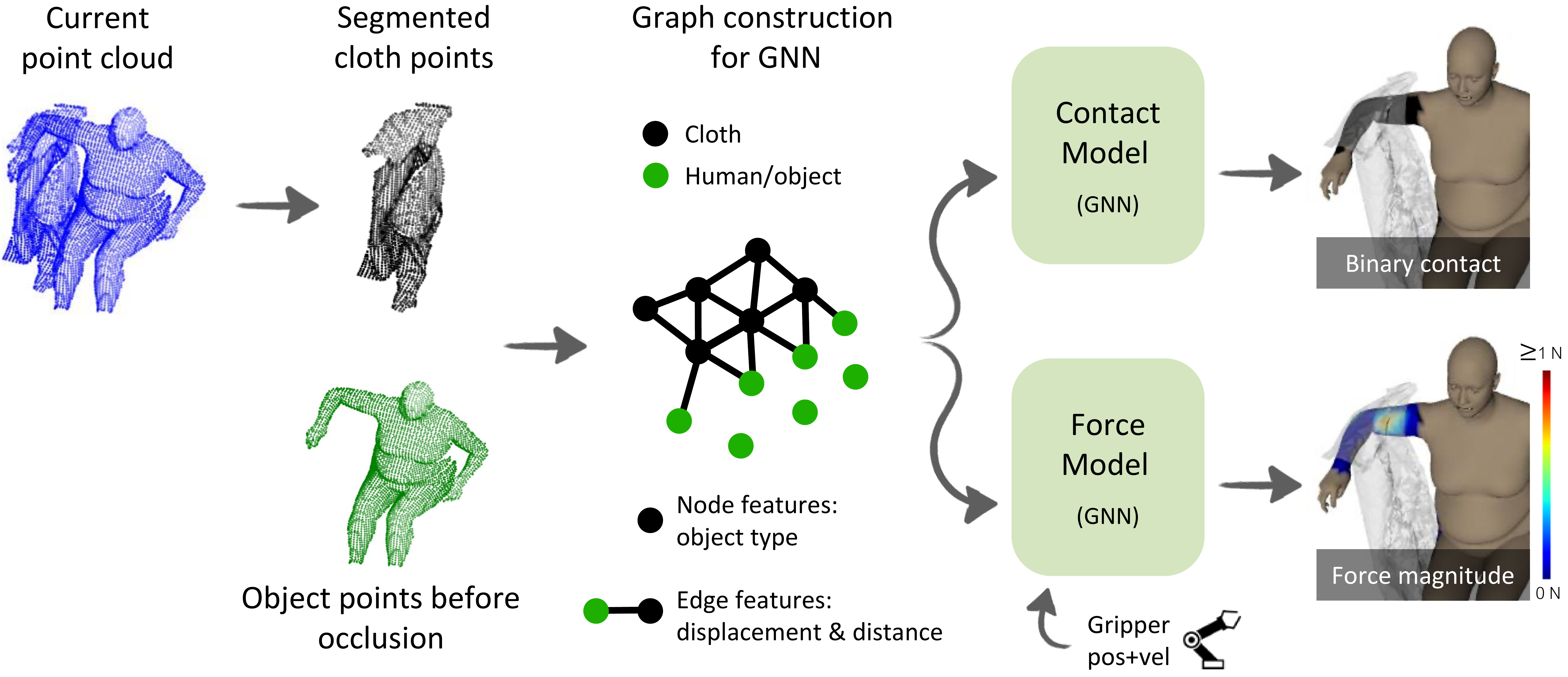}
    \vspace{-0.3cm}
    \caption{
    An overview of the proposed visual haptic reasoning model, which estimates the location and magnitude of applied normal forces when deformable cloth (hospital gown in this example) physically interacts with other objects (human body in this example). The contact prediction model predicts if a point on the human body is in contact with the gown, denoted as the black regions. The force prediction model predicts the magnitude of applied force. Both models take the \yufei{segmented} point cloud as the input, and the force model also takes as input the gripper kinematics. \yufei{In this example, the haptic reasoning models are instantiated as GNNs, and the middle part shows the graph construction process for GNNs.}
    }  
    \label{fig:system}
    \vspace{-0.5cm}
\end{figure*}

\vspace{-0.1cm}
\section{Problem Formulation}
Given a robot manipulation scenario where deformable cloth interacts with another fixed object (e.g., helping a person dress a garment), we aim to learn a model that predicts the applied normal forces between the cloth and the object based on a \yufei{point cloud observation} of the scene.
Formally, given a depth image $I \in \R^{H\times W}$ (where $H$ and $W$ are the height and width of the image) of the scene, we transform the depth image to a point cloud $\bm{P} \in \R^{H\times W \times 3}$ using the camera's intrinsic matrix. 
We then segment the point cloud to obtain a set of points associated with the cloth $\bm{P}_C \subseteq \bm{P}$ and a set of object points $\bm{P}_O \subseteq \bm{P}$. 
\yufei{In simulation, the classification and segmentation of points is provided directly by the simulator. In the real world, we use color thresholding on the RGB image of the scene to perform the segmentation. }
We aim to learn a haptic reasoning model that can predict the magnitude of the contact normal force \yufei{$f_i\in\mathbb{R}$} that cloth applies to each point $\bm{p_i} \in \bm{P}_O$ on the object.

\yufei{
We make the following assumptions for this problem. First, we assume the object remains static during the interaction.
This assumption helps address the visual occlusion of the object caused by cloth during the interaction, as with this assumption we can obtain the object point cloud $\bm{P}_O$ before the cloth occludes the object. Future work can investigate methods such as capacitive sensing~\cite{erickson2021characterizing, erickson2018deep} to track the pose of an object or human limb under visual occlusion.
In addition, we assume the critical contact areas between cloth and an object are observable through a partial point cloud. Future work could incorporate 3D model inference~\cite{mescheder2019occupancy, chi2021garmentnets} and force/torque sensing to predict force distributions over the full 3D model of an object.}

\vspace{-0.1cm}
\section{Cloth Modeling and Contact Force Computation via Physics-based Simulation}
\label{sec:simulation}
\yufei{
We use physics-based simulation to compute the contact force between cloth and other objects, which are used as the ground-truth labels for training our proposed visual haptic reasoning models. Specifically, we choose NVIDIA FleX as our simulator, which uses position-based dynamics~\cite{muller2007position} for cloth simulation. We describe briefly here how position-based dynamics models cloth and how contact forces are computed, and refer the readers to~\cite{muller2007position, macklin2014unified, macklin2019non} for full details. We also note that the proposed framework is orthogonal to the simulation techniques used to compute the contact forces. }

\yufei{
In position-based dynamics, objects are represented using particles and constraints between them. A cloth can be created from a triangular mesh. A particle is created for each vertex in the mesh, and a stretching constraint is created for each edge~\cite{muller2007position, macklin2014unified}, which models a spring that tries to maintain its rest length. 
A bending constraint and a shearing constraint are created for pairs of adjacent triangles. 
Three stiffness parameters $k_{stretch}$, $k_{bending}$ and $k_{share}$ are used to model how strong these constraints are. At each simulation step, after the positions of the particles are updated due to external forces such as gravity, the particle positions are projected  to obey the constraints using a Guassian-Sidel algorithm. }

\yufei{
Each particle in the cloth can have contacts with other objects, e.g., represented using another triangular mesh. The contact forces are solved using the classic Lagrangian dynamics,
with a contact constraint in the form: 
$
c(\bm{p}) = \bm{n}^T(\bm{p} - \bm{q}) - d \geq 0
\label{eq:contact_constraint}
$,
where $\bm{p}$ is the position of the particle, $\bm{n}$ is the normal for the contact plane, and $\bm{q}$ is the point that the particle should not penetrate into, e.g., the vertex of the object mesh. $d$ is a threshold parameter that the particle should maintain from the object point. After the dynamics is jointly solved with the constraint, the contact normal force is then $\bm{f} = \lambda\bm{n}$, and the Lagrangian multiplier $\lambda$ represents the contact normal force magnitude. 
}

\yufei{We now describe how to get the magnitude of the contact normal force $f_i$ for each point on the object $\bm{p_i} \in \bm{P_O}$, which are used as training labels for the proposed visual haptic reasoning models.} In simulation, contact points between cloth and an object may not directly align with object points $\bm{P}_O$ from the observed point cloud.
Given a contact between the cloth and an object that occurs at position $\bm{x}^C_j \in \mathbb{R}^3$ ($j=1,...,N$) \yufei{with normal force} $\bm{f}^C_j \in \mathbb{R}^3$, we distribute the applied force  to all nearby observed object points that are within a distance $\epsilon$ of the contact position. 
This distribution of the contact force is inversely proportionate to the distance between $\bm{x}^C_j$ and $\bm{p}_i$.
Formally, let $\bm{x}_i \in \mathbb{R}^3$ denotes the position of point $\bm{p}_i  \in \bm{P}_O$, 
the contact normal force magnitude $F^C_j = ||\bm{f}^C_j||_2$ is distributed to a point $\bm{p}_i$ as follows:
\begin{equation}
\begin{split}
    F_{j \rightarrow i} &= F_j \cdot \frac{w_i}{\sum_{k=1}^{H\times W} w_k}, \\
    w_i &= \left\{
    \begin{array}{cr}
         \frac{1}{||\bm{x}^C_j - \bm{x}_i||^\xi} &\text{if} ||\bm{x}^C_j - \bm{x}_i||_2 < \epsilon  \\
         0 & Otherwise
    \end{array}
    \right.,
\end{split}
\label{eq:force_distribution}
\end{equation} 
where $\xi > 0$ controls the smoothness of the distribution, which we set to be $0.5$. The contact normal force magnitude $f_i$ on point $\bm{p}_i$ is then the sum of all distributed forces: $f_i = \sum_{j=1}^N F_{j \rightarrow i}$. Our goal is to learn a model $h(\bm{P})$ that takes as input the point cloud $\bm{P}$, and predicts the force magnitude $f_i$ for every point $\bm{p}_i \in \bm{P}_O$ on the object.

\vspace{-0.1cm}
\section{Visual Haptic Reasoning Model}
\vspace{-0.1cm}

\label{sec:method}
\subsection{Method Overview}
\label{sec:overview}
An overview of the \yufei{visual} haptic reasoning framework is shown in Fig.~\ref{fig:system}. 
As the bottom branch shows, we train a force prediction model $h_{force}$, which takes the point cloud $\bm{P}$ \yufei{(consisting of the object points $\bm{P}_O$ and cloth points $\bm{P}_C$)}, and the robot gripper kinematics as input, and predicts the magnitude of the contact force $f_i$ at each point $\bm{p_i} \in \bm{P_O}$.
\yufei{Since the force prediction model is performing regression, it is common to predict small non-zero forces at many points on an object that are not in contact with cloth. To address this issue, we further introduce a contact prediction model $h_{contact}$, shown in the top branch of Fig.~\ref{fig:system}, to predict if a point $\bm{p} \in \bm{P}_O$ is having contact with cloth during manipulation, which removes the need for choosing a force threshold for determining contact at inference time. For training the contact model, a point $\bm{p}$ has a label of 1 if $f_i$ is greater than 0, and a label of 0 otherwise. }

The force prediction model $h_{force}$ and the contact prediction model $h_{contact}$ can be instantiated using different neural network architectures \yufei{(Fig.~\ref{fig:system} shows a GNN instantiation)}. Several learning methods have been designed specifically for modeling point cloud data~\cite{qi2017pointnet, qi2017pointnet++}. In this paper we explore two distinct architectures for the haptic reasoning models, PointNet++~\cite{qi2017pointnet++} and graph neural networks (GNN) ~\cite{lin2022learning, sanchez2020learning}.
In the following subsections, 
we describe in detail how we instantiate the force and contact prediction models as PointNet++ or GNN.

\vspace{-0.1cm}
\subsection{Point Cloud Based Input Representation}
\label{sec:formulation}

We first describe how we construct the input to the PointNet++ and GNN haptic reasoning models based on the point cloud $\bm{P}$, \yufei{which consists of the segmented cloth points $\bm{P}_C$, and object points $\bm{P}_O$ obtained before visual occlusion as described in Section~\ref{sec:formulation}}. 
\yufei{We first filter the point cloud with a voxel grid filter by overlaying a 3D voxel grid over the observed point cloud and take the centroid of the points inside each voxel. This voxelization step is done both in simulation and in the real world, which makes the model agnostic to the density of the observed point cloud, and more robust to sim2real transfer.}
\yufei{Then, given the voxelized point cloud $\bm{P}$ (we overload the notation $\bm{P}$ to refer to the voxelized point cloud in the rest of the paper)}, we remove object points from $\bm{P}_O$ that are sufficiently far from the cloth. We define this new point cloud $\tilde{\bm{P}}$ as:
\begin{equation}
\begin{split}
    \tilde{\bm{P}}&= \tilde{\bm{P}}_O  \cup 
    \{\bm{p}_{gripper}\} \cup  \bm{P}_C,  \\
    \tilde{\bm{P}}_O &= \{\bm{p}_i | (\exists\bm{p}_i \in \bm{P}_O) (\exists\bm{p}_j \in \bm{P}_C) [||\bm{x}_i - \bm{x}_j||_2 < \tau]\},
\end{split}
\label{eq:filter_eq}
\end{equation}
where $\bm{x}_i$ represents the position of point $\bm{p}_i$, and $\tilde{\bm{P}}_O$ denotes object points that are within a distance $\tau$ of some cloth point.
 $\bm{p}_{gripper}$ is a single point at the position of the robot gripper.

$\tilde{\bm{P}}$ can be directly used as input to a PointNet++ model. 
To form a GNN model, we build a graph $G = \langle V, E \rangle$ from the point cloud, where $V$ are the nodes and $E$ are the edges of the graph. \yufei{Fig.~\ref{fig:system} visualizes the graph construction process.}
The nodes $V$ of the graph simply consist of all the
points in $\tilde{\bm{P}}$. 
For edges, we connect an edge $e_{jk}$ between a node $j$ and node $k$ when the following criteria are satisfied: 1) the distance between nodes is below a threshold $\alpha$, i.e. $||\bm{x}_j - \bm{x}_k||_2 < \alpha$, where $\bm{x}_j$ denotes the position of node $j$, 
and 2) at least one node is a point on the cloth.
Intuitively, message passing along the edges of the GNN can simulate the propagation of force from the gripper to the cloth, within the cloth itself, and from the cloth to the object. 

\vspace{-0.15cm}
\subsection{Force Prediction Model}
\vspace{-0.1cm}
\label{sec:force}

We describe two different instantiations for the force prediction model $h_{force}$, which is capable of predicting the forces that cloth applies onto objects in contact. The first instantiation uses a PointNet++, which can be directly applied to the cropped point cloud $\tilde{\bm{P}}$. The input includes the position $\bm{x}$ of each point $\bm{p} \in \tilde{\bm{P}}$ \yufei{and a feature vector associated with each point $\bm{p}$. The feature vector consists of robot gripper velocity $\bm{v}$} and a $3$-dimensional one-hot encoding vector $[\mathbf{1}_{object}(\bm{p}), \mathbf{1}_{cloth}(\bm{p}), \mathbf{1}_{gripper}(\bm{p})]$ indicating the type of the point (where $\mathbf{1}_{y}(\bm{p})$ is $1$ if $\bm{p}$ belongs to $y$ and $0$ otherwise). We normalize the positions of the point to be zero-mean so that the model is invariant to translations of the point cloud. 

To model force predictions with a GNN, the input to the force model is the graph built from the point cloud, as described in Section~\ref{sec:formulation}. We use the GNS architecture~\cite{sanchez2020learning,lin2022learning} for the GNN model.
The features for each node in the graph include the one-hot encoding vector $[\mathbf{1}_{object}(\bm{p}), \mathbf{1}_{cloth}(\bm{p}), \mathbf{1}_{gripper}(\bm{p})]$ and the gripper velocity $\bm{v}$. The edge feature of an edge connecting node $i$ and node $j$ includes the distance vector $\bm{x}_i - \bm{x}_j$ and its L2 norm $||\bm{x}_i - \bm{x}_j||_2$. By using only relative distance in the edge features, the GNN model is also invariant to translations of the point cloud in Cartesian space. We do not include positions in the node features since the edge features are sufficient to capture the relative relationship between nodes.

As described in Section~\ref{sec:overview},
the output of the force prediction model $h_{force}$ 
is the estimated magnitude of the contact normal force $\hat{f}_i$ for every point $\bm{p_i} \in \bm{\tilde{P}_O}$. 
We train the force model using a Mean Squared Error loss between the predicted contact force 
$\hat{f_i}$
and the ground-truth contact force $f_i$, which can be obtained in simulation and computed using Eq~\eqref{eq:force_distribution}. 
We mask the loss to be computed only on points that are having contact during the interaction, i.e., for points whose contact force $f_i$ is greater than $0$.

\vspace{-0.15cm}
\subsection{Contact Prediction Model} 
\vspace{-0.1cm}
\label{sec:contact}

When training both PointNet++ and GNN models for contact prediction, we remove the gripper point $\bm{p}_{gripper}$ from the point cloud and reduce the point/node features to include only a $1$-dim one-hot encoding $\mathbf{I}_{object}(p)$. A PointNet++ for contact prediction still includes normalized point positions as part of its input, and a GNN for contact prediction uses the same edge features as presented in Section~\ref{sec:force}.

The output of the contact model $h_{contact}$
is the probability $m_i$ of each point $\bm{p}_i \in \bm{\tilde{P}_O}$ being in contact with the cloth. We train it with a Binary Classification Loss, where the ground-truth contact information can be obtained in simulation. For each point $\bm{p}_i \in \tilde{\bm{P}}_O$, the ground-truth label is $1$ if it has non-zero contact force $f_i$ (computed using Eq~\eqref{eq:force_distribution}), otherwise the label is $0$.  
At test time, we combine the predictions of the contact and the force prediction model to compute the final contact normal force magnitude of a point $\bm{p}_i$: 
\begin{equation}
    f^{pred}_i =  \left\{
    \begin{array}{cc}
    \hat{f}_i     & \text{if}~m_i > \beta \\
    0     & \text{Otherwise}
    \end{array}
    \right.,
    \label{eq:predict}
\end{equation}
where $\beta$ is a decision threshold.

\section{Tasks and Dataset Collection}

\begin{figure}[t]
    \centering
    \includegraphics[width=0.2\columnwidth]{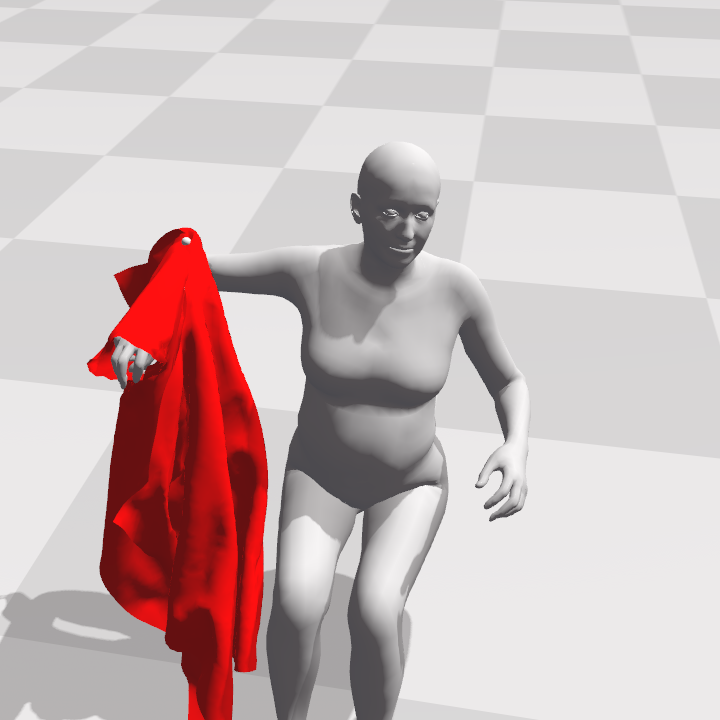}
    \includegraphics[width=0.2\columnwidth]{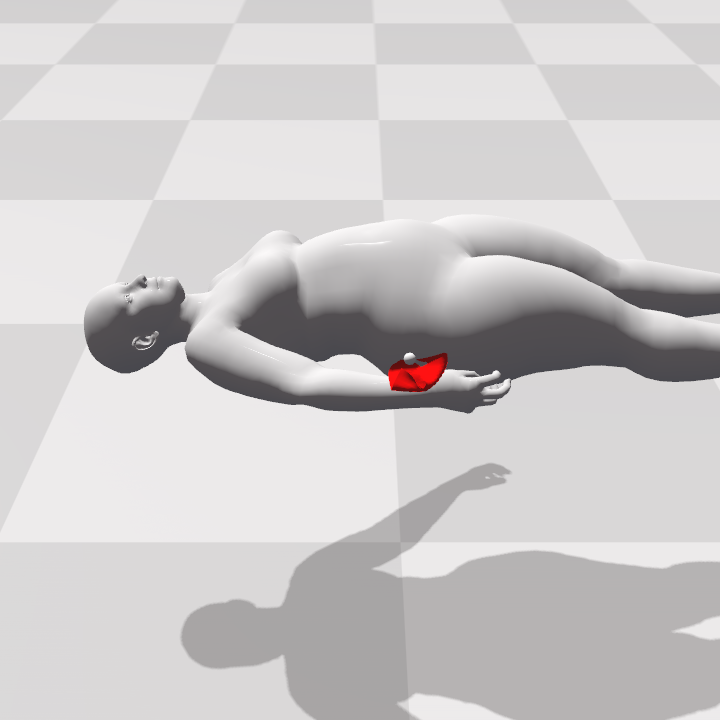}
    \includegraphics[width=0.2\columnwidth, trim={8cm 8cm 8cm 8cm}, clip]{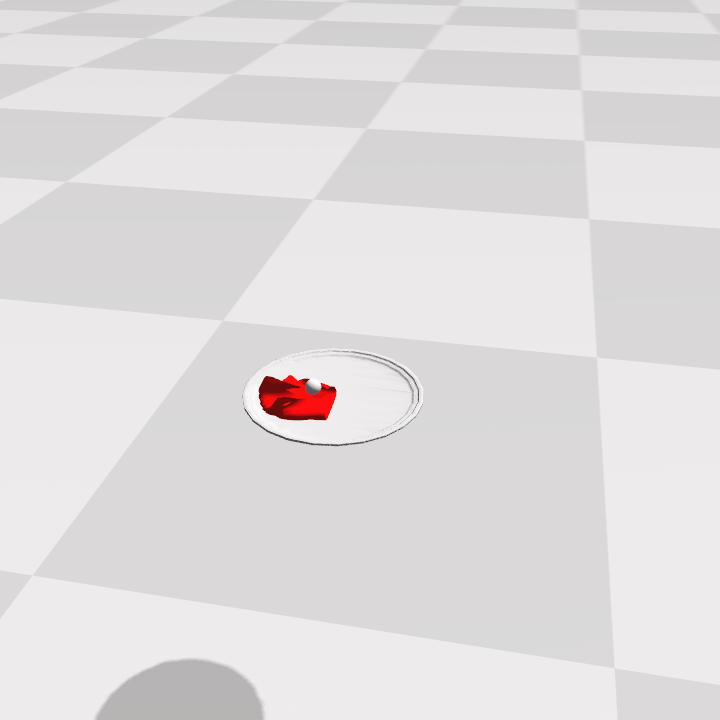} \includegraphics[width=0.2\columnwidth]{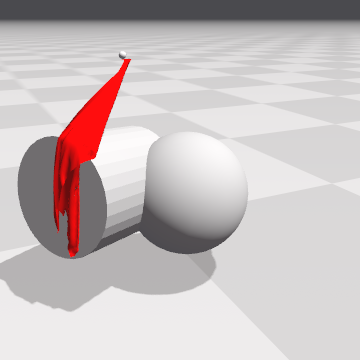}
 \caption{Visuals of the 4 simulation tasks. From left to right: assistive dressing, assistive bathing, dish washing, and primitive shapes. }
    \vspace{-0.7cm}
    \label{fig:simulation_task}
\end{figure}

\yufei{As described in Sec.~\ref{sec:simulation},
we use the NVIDIA FleX wrapped in SoftGym~\cite{Lin-2020-127232} as our physics simulation.}
As shown in Fig.~\ref{fig:simulation_task},
we build three representative robotic assistive and manipulation tasks that involve physical interactions between cloth and other objects: 1) Assistive dressing, where a robot must dress a hospital gown onto a person's right arm; 2) Assistive bathing, where the robot manipulates a washcloth to clean a person's arm; 3) Dish washing, where the robot uses a washcloth to clean the surface of a dish. 
To further evaluate visual haptic reasoning between cloth interacting with arbitrary rigid objects, we build a task where the robot pulls a rectangular towel over random combinations of objects with primitive shapes, including cylinders, cubes and spheres (referred to as the primitive shape task). 
In all tasks, we use a point grasp (visualized as a small white sphere), and the grasping is simulated by creating rigid anchors between the point grasp and the closest particles on the cloth. 

We add the following variations for these tasks. For assistive dressing and bathing, we use the SMPL-X~\cite{pavlakos2019expressive} model to generate 100 human body meshes varying in body shape and size. For dish washing, we randomly select 3 plates from the ShapeNet  dataset~\cite{chang2015shapenet}. For primitive shape, we vary the number of primitive shapes between 1 to 3, use a random selection of chosen shapes (cylinder, sphere, or cube), and vary the size and pose of each shape. 
For all tasks, we generate the gripper movement trajectories by linearly interpolating between some way points, where the way points are chosen differently for each task. For the dressing and bathing task, the waypoints consist of the  fingertip, wrist, elbow, and shoulder of the human arm, with random deviation uniformly sampled from $[-5, 5]$cm added to these waypoints. For dish washing, we randomly sample way points on the surface of the plate. For primitive shape, we randomly sample waypoints on the surface contour of the combination of the primitive objects.
For all tasks, we vary the gripper movement velocity across different trajectories. 
Depending on the gripper velocity and the waypoint locations, each simulated cloth manipulation trajectory can have between $200$ to $800$ time steps, where at each time step we store the \yufei{point cloud}, gripper velocity, and ground-truth contact normal forces between the cloth and the object as a data point for training.  
\begin{table}[t]
    \centering    \scriptsize
    \begin{tabular}{c|c|c|c|c}

    \toprule
      &  \specialcell{Assistive \\ Dressing} & \specialcell{Assistive \\ Bating} & \specialcell{Dish \\ Washing} & \specialcell{Primitive \\ Shapes}  \\ \hline
    Training size &  81379 & 187568  & 258624 &  624724 \\ \hline
     Validation size  & 7948  &  17978 & 24003 &  69413 \\ \hline
Test size & 3977  & 8989 & 11853 & 33889 \\ \hline
average \# cloth points & 2361  & 79 & 76 & 353 \\ \hline
average \# object points & 267  & 114 & 130 & 370 \\ 
    \bottomrule
    \end{tabular}
    \vspace{0.2cm}
    \caption{Statistics of the collected dataset.}
    \vspace{-0.8cm}
    \label{tab:dataset_statistics}
\end{table}
For assistive dressing, assistive bathing, and dish washing, we collect 600, 60, and 30 trajectories for training, validation and test, while due to the large task variation of primitive shape task, we collect 1800, 200, and 90 trajectories for training, validation, and test.
The exact number of data points for the collected datasets is summarized in Table~\ref{tab:dataset_statistics}. 
\yufei{The average number of object points and cloth points that will be inputted to our model, i.e., size of $\bm{\tilde{P}}_O$ and $\bm{P}_C$, are also reported in Table~\ref{tab:dataset_statistics}. }

\vspace{-0.05cm}
\section{Experimental Results}
\vspace{-0.05cm}
\subsection{Evaluation Metrics and Baselines}
For evaluation metrics, we use Mean Absolute Error (MAE) \yufei{in Newtons} for force prediction in relation to the ground-truth contact forces, and F1 score for contact prediction. 
For all learning methods, we search over the contact prediction decision threshold ($\beta$ in Eq.~\eqref{eq:predict}) on the validation dataset, and use the threshold that produces the highest F1 score. 

We compare the proposed GNN and PointNet++ learning methods with the following baselines:
\textit{MLP}, where we trained two separate Multilayer Perceptrons that take as input a vectorized point cloud, one for contact prediction and one for force prediction. As the number of points in the point cloud can vary across tasks and trajectories, we take the maximum number of points as the fixed input dimension and pad the observation with 0 when it has fewer dimensions.
\textit{Constant Force Prediction}, a force prediction baseline which predicts a constant force for every point in contact. The constant force is chosen to be the median force on the training dataset which gives the lowest MAE on the training dataset. 
\textit{Neighborhood Contact Prediction}, a contact prediction baseline that predicts a point on the object to be in contact if its distance to the closest point in the cloth is below a threshold. We perform a grid search over the threshold on the training dataset and evaluate with the one that results in the best F1 score.

For GNN, PointNet++, and MLP, we train two variants: the first is the task-specific model that is trained with data only from a single task, and the second is the task-agnostic model that is trained with data from all three assistive dressing, assistive bathing, and dish washing tasks. We hold out the primitive shape task and use it to evaluate the generalization performance of our task-agnostic models.
In all evaluation tables, we use the suffix `-S' to denote the task-specific models, the suffix `-A' to denote the task-agnostic models, and we use bold text to denote the best result, and underlined text to denote the second best result.

\vspace{-0.1cm}
\subsection{Implementation Details}

\yufei{For PointNet++, we use the standard segmentation type of architecture in the original paper~\cite{qi2017pointnet++}.
 }
\yufei{
For GNN, we use the standard GNS~\cite{sanchez2020learning} architecture. More details about the network architectures can be found on our project website.
}
\yufei{For the MLP baseline, we use a MLP of $[1024, 512, 256, 512, 1024]$ neurons and ReLU activation.}
\yufei{
We train the PointNet++, GNN and MLP models using the Adam~\cite{kingma2014adam} optimizer, with a learning rate of $0.0001$ and batch size of 8. We train all models until they converge on the training dataset, and pick the model that has the lowest loss on the validation dataset for evaluating on the test dataset.}
\yufei{For NVIDIA FleX simulator, the particle radius in the simulator is $0.625$cm. 
Due to different sizes of cloth used in different tasks, we tune the stiffness of the stretch, bending, and shear constraints to ensure stable simulation behaviours of cloth for different tasks. The stiffness of these constraints are set to be $[1.7, 1.7, 1.7]$ for assistive dressing, and $[1.0, 0.9, 0.8]$ for the other three tasks. We set the threshold for contact ($d$ in Sec.~\ref{sec:simulation}) to be $5$mm.}
\yufei{For hyper-parameters described in Sec.\ref{sec:method}, the value of $\epsilon$ used in Eq.\eqref{eq:force_distribution} is set to be $3.12$cm, the voxel size we used to voxelize the point cloud is $1.56$cm, the value of $\tau$ in Eq.\eqref{eq:filter_eq} is $6.25$cm, and the value of $\alpha$ for determining the edge connection for GNN is $3.75$cm.}

\begin{table}[t]
    \centering    \scriptsize
    \begin{tabular}{c|c|c|c|c}
    \toprule
      \diagbox{Method}{Task}&  \specialcell{Assistive \\ Dressing} & \specialcell{Assistive \\ Bating} & \specialcell{Dish \\ Washing} & \specialcell{Primitive \\ Shapes}  \\ \hline
    GNN-S & 0.200 & \textbf{0.028} & \underline{0.064}  & \textbf{0.931}  \\ \hline
    GNN-A & 0.192  & 0.035 & \textbf{0.063} & 1.460  \\ \hline
PointNet++-S & \textbf{0.188} & \underline{0.029}  & 0.066  & \underline{1.073}   \\ \hline
PointNet++-A & \underline{0.189}  & 0.034 & 0.065  &  1.398\\ \hline
MLP-S & 0.640  & 0.057   & 0.134  & 1.503  \\ \hline
MLP-A & 0.696 & 0.057  &0.122  & 1.502 \\ \hline
Constant Force & 0.555 & 0.059 & 0.118 & 1.367 \\

    \bottomrule
    \end{tabular}
    \vspace{0.2cm}
    \caption{Force prediction MAE \yufei{(in Newtons)} on test dataset.}
    \vspace{-0.8cm}
    \label{tab:force_performance}
\end{table}

\vspace{-0.25cm}
\subsection{Simulation Results}
\subsubsection{Force Prediction Result}
The force prediction MAE of all methods over all tasks in simulation are shown in Table~\ref{tab:force_performance}. 
As shown in the first 3 columns, for assistive dressing, assistive bathing, and dish washing, for either task-agnositic or task-specific models, both GNN and PointNet++ models achieved significantly lower error than the constant force prediction baseline and the MLP baseline. 
Both GNN and PointNet++ models achieved similar performance overall, with GNN performing slightly better on bathing and dish washing, while PointNet++ achieved lower error for the assistive dressing task.
Interestingly, for both GNN and PointNet++, the task-agnostic models achieved similar prediction errors compared with the task-specific models, indicating we can train a single model on multiple tasks instead of one for each task. 
When compared to task-specific models trained on primitive shapes, we observe from the $4^{th}$ column of Table~\ref{tab:force_performance} that task-agnostic force prediction models do not generalize as well to the primitive shape task, due to the large distribution shift.
In addition, a task-specific GNN performs better than task-specific PointNet++ for inferring applied forces during the primitive shapes task, indicating that GNN architectures may be better suited for tasks with very large variations.  
\yufei{We also report the percentage of error, i.e., the ratio between the force prediction MAE and the mean value of the ground-truth force of the dataset, for the best methods on each dataset. For assistive dressing, the PointNet++-S has an error percentage of $33.1\%$, for assistive bathing, the GNN-S has an error percentage of $34.0\%$, for dish washing, the GNN-A model has an error percentage of $43.2\%$, and for primitive shapes, the GNN-S model has an error percentage of $68\%$ due to the large variation for this task. We later show in Fig.~\ref{fig:plate-clean} that the force predictions are good enough for potential downstream tasks.}
\yufei{On a NVIDIA 3090 GPU, the inference time for the GNN force model is $\sim$10ms, and the inference time for the PointNet++ force model is $\sim$20ms. }

\begin{table}[t]
    \centering    \scriptsize
    \begin{tabular}{c|c|c|c|c}

    \toprule
     \diagbox{Algorithm}{Task} &  \specialcell{Assistive \\ Dressing} & \specialcell{Assistive \\ Bating} & \specialcell{Dish \\ Washing} & \specialcell{Primitive \\ Shapes}  \\ \hline
    GNN-S &  0.888 & 0.910  & \underline{0.955} &  \underline{0.930} \\ \hline
     GNN-A & 0.890  &  0.912 & 0.954 &  0.872 \\ \hline
PointNet++-S &\textbf{ 0.933} & \textbf{0.953} & \textbf{0.956} & \textbf{0.949 } \\ \hline
PointNet++-A & \underline{0.930} & \underline{0.946} & 0.948 &  0.866 \\ \hline
MLP-S & 0.790 & 0.748 & 0.902 & 0.781 \\ \hline 
MLP-A & 0.788 & 0.728 & 0.888 & --- \\ \hline 
Neighborhood & 0.740 & 0.844 & 0.873 & 0.843 \\ 
    \bottomrule
    \end{tabular}
    \vspace{0.2cm}
    \caption{Contact prediction F1 on  test dataset.}
    \vspace{-0.4cm}
    \label{tab:contact_performance}
\end{table}

\subsubsection{Contact Prediction Result}
Table~\ref{tab:contact_performance} shows the contact prediction F1 score of different methods on all tasks in simulation. For all three assistive tasks shown in the first 3 columns, both GNN and PointNet++ achieved substantially higher F1 scores than the Neighborhood Contact Prediction and the MLP baselines. For contact prediction, PointNet++ is consistently better than GNN, for both task-agnostic and task-specific models. 
We also observe similar F1 scores for both task-agnostic and task-specific contact prediction models, indicating that we are able to train a single model on multiple tasks.
In contrast to force prediction, the task-agnostic GNN and PointNet++ models generalized well to predicting cloth-object contact for the primitive shape task.
The task-agnostic MLP model is unable to generalize due to its limiting way of handling point cloud data.
\yufei{The inference time for the GNN contact model is $\sim$10ms, and the inference time for the PointNet++ contact model is $\sim$20ms. }

\begin{figure*}[t]
    \centering
    \includegraphics[width=.75\textwidth]{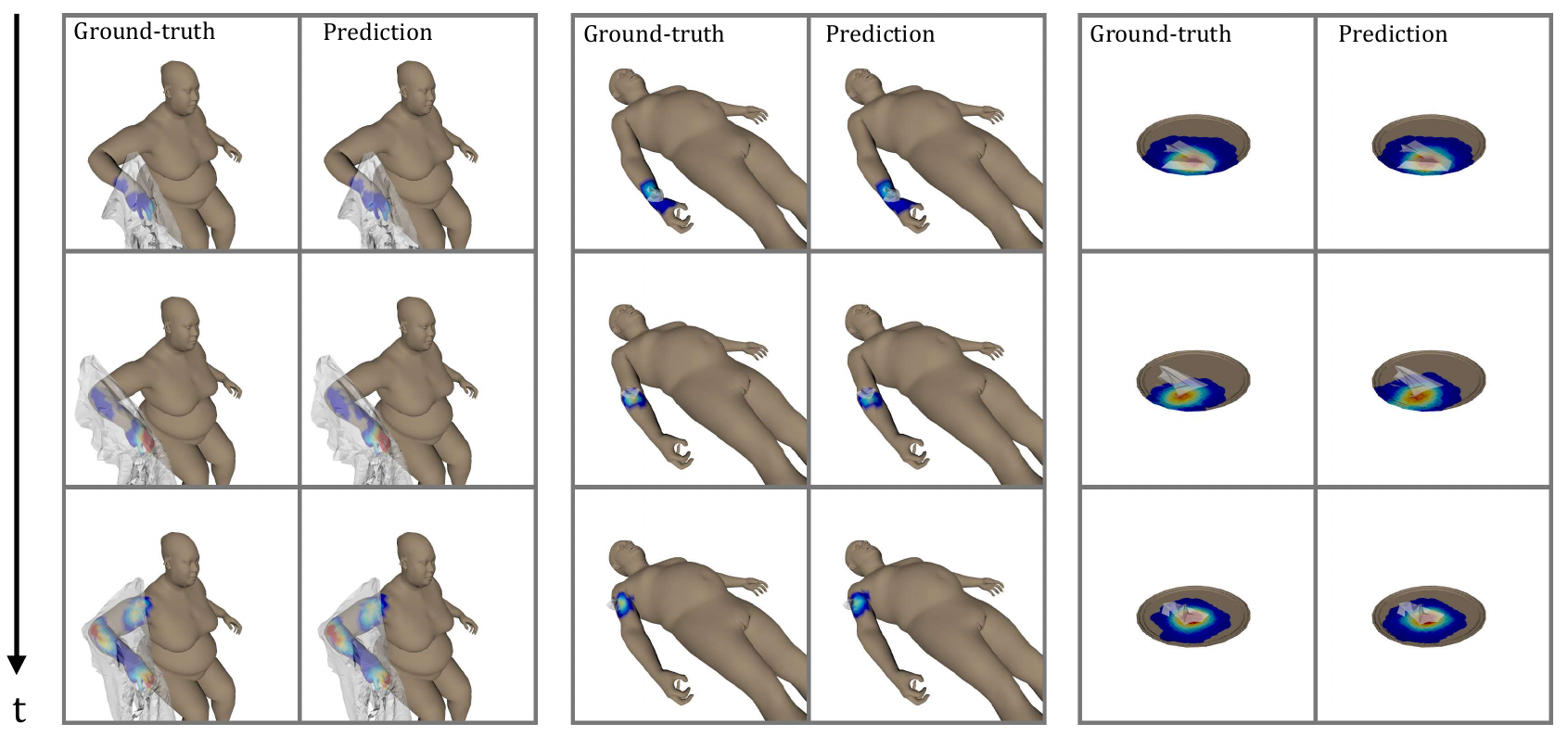}
    \vspace{-0.4cm}
    \caption{Visuals of the force and contact predictions of the task-agnostic PointNet++ model for the assistive dressing, assistive bathing, and dish washing tasks in simulation, on the test dataset. For each task, the \yufei{left column} is the ground-truth and the \yufei{right column} is the prediction. \yufei{The color map is scaled differently for each task due to different magnitudes of forces, but the same color map is used to visualize the ground-truth and predicted forces within each task. Better viewed zoomed-in.}
    }
    \vspace{-0.3cm}
    \label{fig:simulation_visual}
\end{figure*}

\subsubsection{Visualizations of Contact and Force Predictions}
Fig.~\ref{fig:simulation_visual} compares the predicted contact and force distributions from the task-agnostic PointNet++ model with the ground-truth for assistive dressing, assistive bathing, and dish washing tasks in simulation. As shown, the predicted contact areas and force magnitudes are qualitatively similar to ground-truth. In assistive dressing, the model accurately predicts large forces around the finger, elbow, and shoulder when the gown gets caught at those regions, and in bathing and dish washing, the model accurately predicts a spherical shape force distribution that decays from the center of the contact. More visuals on varying trajectories, human body shapes, and plate sizes can be found on the project webpage.

\begin{table}[t]
    \centering    \scriptsize
    \begin{tabular}{c|c|c|c|c}
    \toprule
     \diagbox{Method}{Task} &  \specialcell{Assistive \\ Dressing} & \specialcell{Assistive \\ Bating} & \specialcell{Dish \\ Washing} & \specialcell{Primitive \\ Shapes}  \\ \hline
    GNN-S & 0.200  & \textbf{0.028}  & \underline{0.064} & \textbf{0.931}  \\ \hline
    GNN-S-particle & \textbf{0.160}  & \textbf{0.026}  & \underline{0.059} & \textbf{0.622} \\ \hline
   PointNet++-S & \textbf{0.188} & \underline{0.029}  & 0.066  & \underline{1.073}    \\ \hline
PointNet++-S-particle & 0.193  & 0.029 & 0.067   & \underline{0.734} \\ \toprule

    GNN-A & 0.192   & 0.035 & \textbf{0.063} & 1.460  \\ \hline
    GNN-A-particle & \underline{0.178}  &  0.030 &  \textbf{0.055} & 1.444 \\ \hline
    PointNet++-A & \underline{0.189}  & 0.034 & 0.065 &  1.398 \\ \hline
PointNet++-A-particle & 0.194  & 0.034 & 0.034  & 1.431 \\ 
    \bottomrule
    \end{tabular}
    \vspace{0.2cm}
    \caption{Force prediction MAE: point cloud vs. cloth particles. }
    \vspace{-0.8cm}
    \label{tab:force_particle}
\end{table}

\begin{table}[t]
    \centering    \scriptsize
    \begin{tabular}{c|c|c|c|c}
    \toprule
     \diagbox{Method}{Task2} &  \specialcell{Assistive \\ Dressing} & \specialcell{Assistive \\ Bating} & \specialcell{Dish \\ Washing} & \specialcell{Primitive \\ Shapes}  \\ \hline
    GNN-S & 0.888  & 0.910 &  0.955 & 0.930 \\ \hline
    GNN-S w/ particles & \underline{0.944}  & \underline{0.961}  & \textbf{0.970} & \textbf{0.973} \\ \hline
PointNet++-S & 0.933 & 0.953 &  0.956 & 0.949 \\  \hline
PointNet++-S w/ particles & 0.932 & 0.950 & 0.954  & \underline{0.961} \\ \toprule

    GNN-A & 0.890  & 0.912 & 0.954  & 0.872 \\ \hline
    GNN-A w/ particles &  \textbf{0.945} &  \textbf{0.962} & \textbf{0.970} &  0.898 \\ \hline
PointNet++-A & 0.930 & 0.946 & 0.948  & 0.866 \\ \hline
PointNet++-A w/ particles & 0.932 & 0.950 &  0.954 & 0.893 \\ 

    \bottomrule
    \end{tabular}
    \vspace{0.2cm}
    \caption{contact prediction F1: point cloud vs. cloth particles.}
    \vspace{-0.8cm}
    \label{tab:contact_particle}
\end{table}

\subsubsection{Comparison to Using Ground-truth Cloth Particles}
We investigate how haptic reasoning models perform when we replace the cloth point cloud with ground-truth cloth particles in FleX simulation. Intuitively, since the cloth particles perfectly represent the underlying cloth mesh and its deformation, we would expect using particles to result in lower force prediction error and higher contact prediction accuracy. The object is still represented using point cloud.

Table~\ref{tab:force_particle} shows the force prediction result with particles, and Table~\ref{tab:contact_particle} shows the contact prediction result. Interestingly, we find that for both force and contact prediction, using particles with GNN models resulted in higher performance, yet using ground-truth particles did not improve performance with PointNet++ models. 
We attribute this to the message passing scheme in GNN, which may leverage the particle representations better compared with the abstraction layers in PointNet++. 
We also note that for both contact and force prediction, among all compared methods and for all tasks, the best-performing method is always a GNN with access to ground-truth cloth particles.
This finding suggests an interesting direction for future work in tracking the underlying mesh particles of deformable cloth in the real world.

\subsubsection{Sensitivity to Noise in Point Cloud}
\yufei{
Since point clouds obtained in the real world are usually noisy, we test the sensitivity of our method to noise in the point cloud in simulation. 
We manually inject different levels of noise to the point cloud positions, which are modeled as Gaussian distributions with different standard deviations (Std)~\cite{ahn2019analysis}. We test the GNN-A model on the assistive dressing task. The force model is robust up $3$mm of noise, where the MAE slightly increases from $0.189$ to $0.253$, while the contact model is robust up to $10$mm of noise, with the contact F1 score drops slightly from $0.889$ to $0.808$. The project website details more experiments on the models' sensitivity to noise.
}

\begin{figure*}[t]
    \centering
    \includegraphics[width=.3\textwidth]{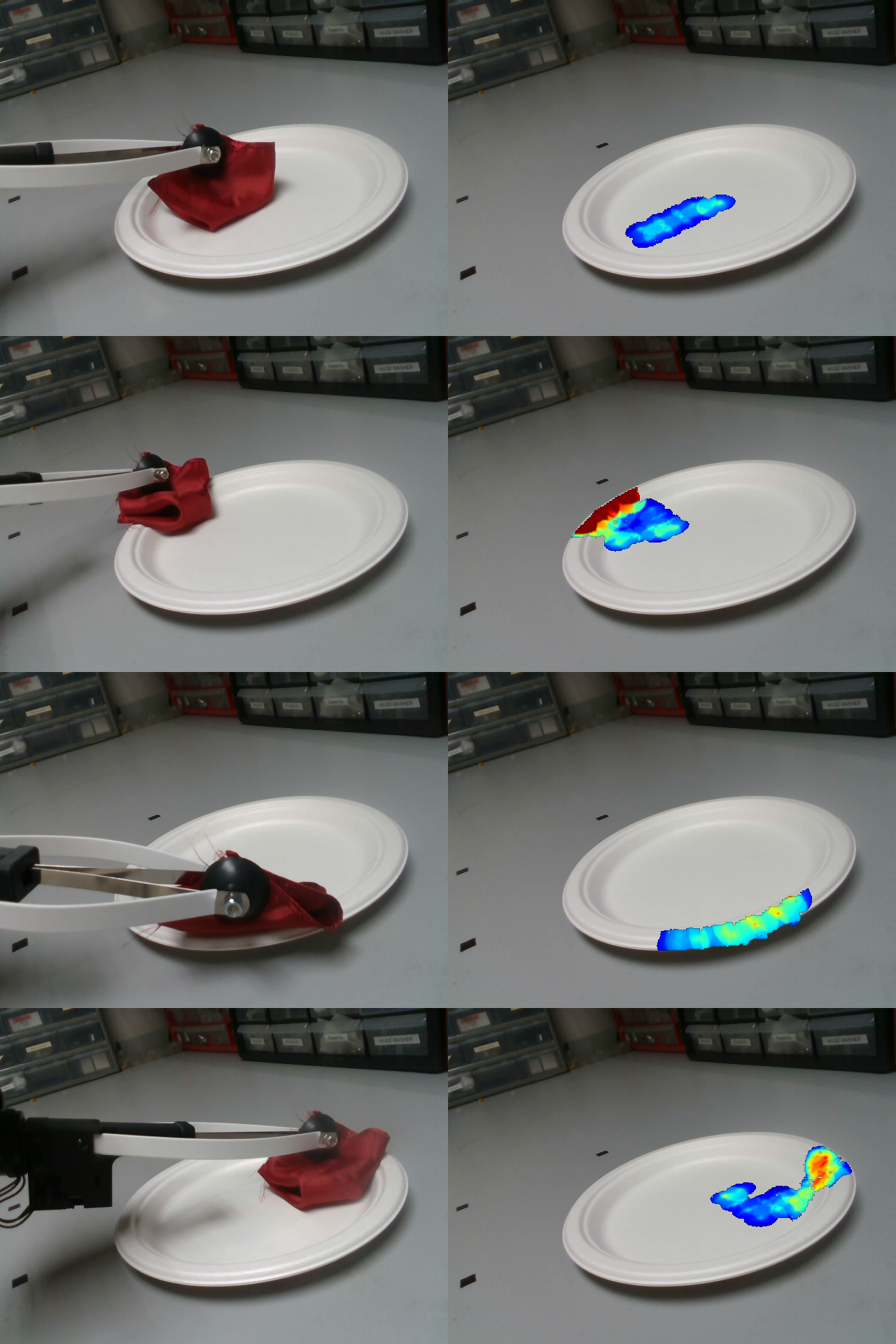}
    \includegraphics[width=.3\textwidth]{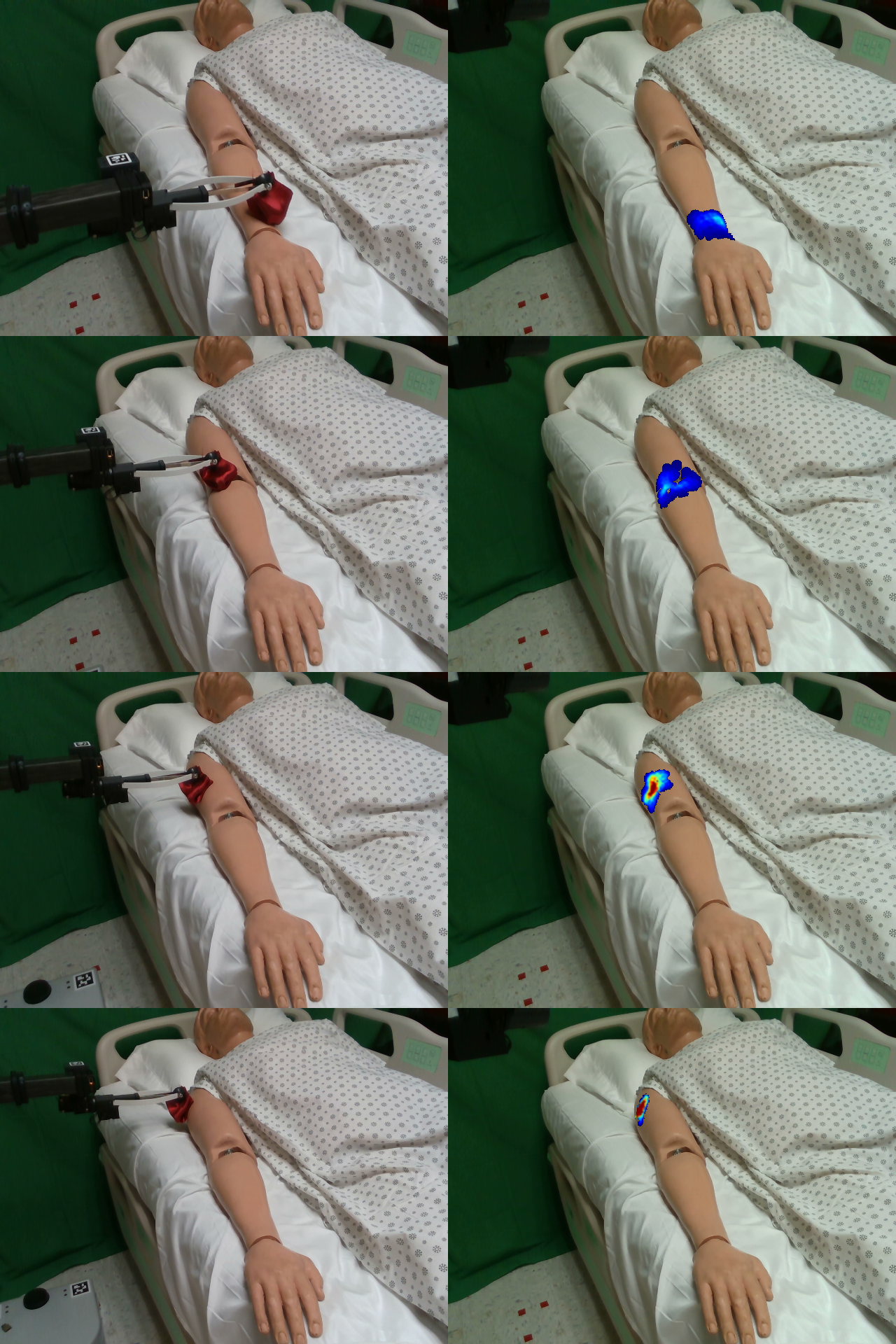}
    \includegraphics[width=.234375\textwidth]{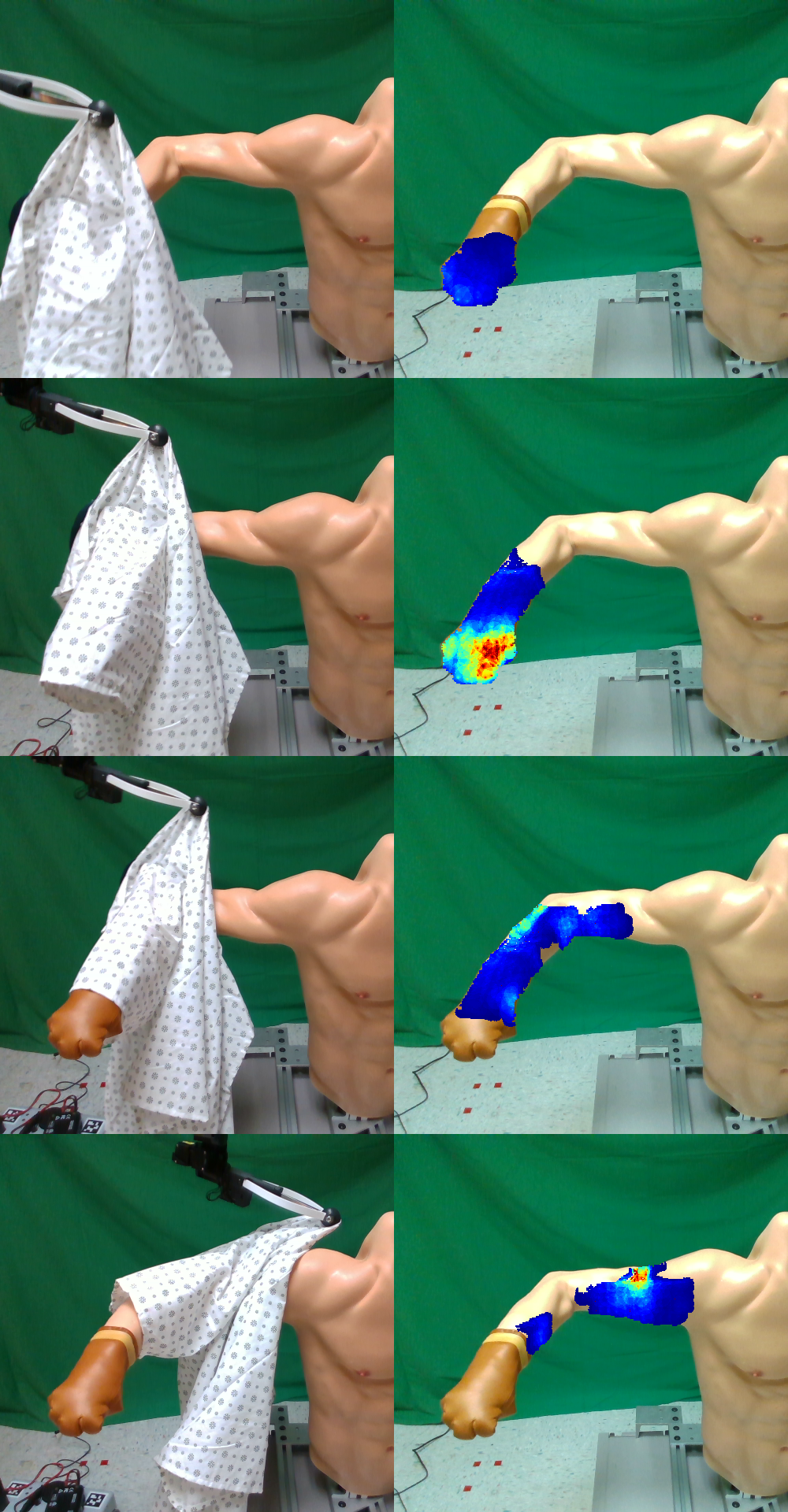}
    \vspace{-0.3cm}
    \caption{
    Qualitative visualizations of the force and contact predictions of the trained visual haptic reasoning models for real-world tasks. 
    Left:  GNN-A for dish washing; Middle: PointNet++-A for bathing a manikin; Right: PointNet++-S for dressing a gown on a silicone arm. All models are trained entirely in simulation. Better viewed zoomed-in. \yufei{The colors are scaled differently for each task to better show different magnitudes of forces in different tasks. }
    }
    \vspace{-0.6cm}
    \label{fig:real_world_visual}
\end{figure*}

\begin{figure}[t]
    \centering
    \includegraphics[width=.8\columnwidth]{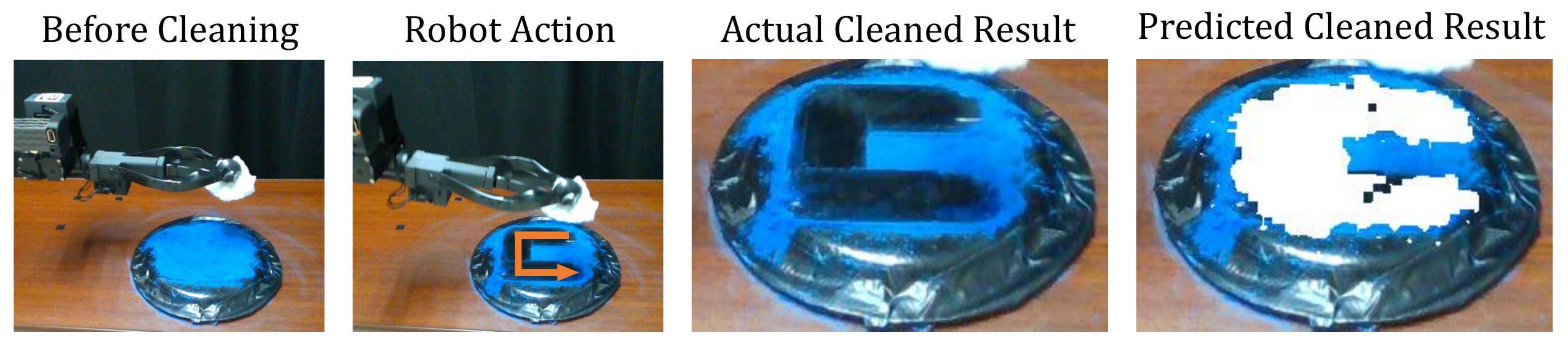}
    \vspace{-0.4cm}
    \caption{\yufei{Comparison of the predicted cleaned area (the white regions in the 4th plot) of the GNN-S model and the actual cleaned area (the 3rd plot) after the robot wipes the blue powder on the black plate with a wash cloth.}}
    \label{fig:plate-clean}
    \vspace{-0.8cm}
\end{figure}

\vspace{-0.2cm}
\subsection{Real World Results}
\subsubsection{Experimental Setup}
We evaluated our model trained purely in simulation on the same three tasks in the real world with a Stretch RE-1 mobile manipulator. For assistive bathing, we use a medical manikin lying on a hospital bed for the human body. For assistive dressing, we use a silicone torso and arm model that is posed similar to human models in simulation. For dish washing, we pick a common dinner plate.
\yufei{Fig.~\ref{fig:real_world_visual} visualizes these three tasks.}
We use the same type of control trajectories as we used in simulation. An Intel D435i camera is used to capture a depth image of the scene, and we use color thresholding on the RGB image to segment the cloth and the object. 
Due to the challenge of sensorizing these environments with high-resolution force sensing arrays, we provide qualitative results of the contact and force predicted by our trained models in these three tasks.
\yufei{We conduct an additional experiment to illustrate the accuracy of the trained models. As shown in Fig.~\ref{fig:plate-clean}, we command the Stretch RE-1 robot to clean the blue powder on a black plate, and we use the trained model to predict which regions are cleaned after the robot action. A point on the plate is assumed to be cleaned if the predicted force magnitude is above a chosen threshold. Such predictions can be potentially used for downstream control tasks to clean specific area of the plate.}

\subsubsection{Qualitative Results}
As shown in Fig.~\ref{fig:real_world_visual}, the various haptic reasoning models trained entirely in simulation produce qualitatively reasonable contact and force distribution predictions for all three tasks. 
As the gripper and washcloth move over the edge of the plate during dish washing, our trained haptic reasoning models (GNN-A) predict a visually accurately force and contact distribution, with greater forces applied on the edges of the plate.
For dressing assistance, we observe the contact model (PointNet++-S) predicts sizable regions of contact between the garment and body that align with visible observations. We observe greater forces on the hand as the hospital gown sleeve is initially pulled onto the hand and briefly snags, and we observe larger force predictions near the shoulder as the robot begins to stretch and pull the garment over the shoulder.
More trajectories \yufei{and some failure cases} are in the supplement video and on the project webpage.
\yufei{As Fig.~\ref{fig:plate-clean} shows, the trained model gives qualitatively correct predictions on which regions of the plate were cleaned by the wash cloth, demonstrating the potential of using visual haptic reasoning models for downstream control tasks. 
All these results indicate that haptic reasoning can be reasonably transferred from simulation to the real world, and presents a promising direction for continued research in robotic caregiving and robotic cloth manipulation.}

\vspace{-0.2cm}
\section{CONCLUSION}
\vspace{-0.1cm}

We introduced a formulation for robots to compute haptic reasoning during cloth manipulation. Haptic reasoning enables a robot the infer the distribution of applied forces as cloth interacts with other objects or people in the environment. We presented two distinct model representations, including GNN and PointNet++ implementations, that enable haptic reasoning using only point cloud and robot kinematic observations. 
We conducted quantitative analyses of model performance during robot-assisted dressing, bathing, and dish washing tasks in simulation and demonstrated generalization performance across human body sizes and object shapes. We also transferred simulation-trained models to a real environment and evaluated visual haptic reasoning with a mobile manipulator performing physically assistive tasks with cloth.

\section*{Acknowledgment}
The authors would like to thank Fukang Liu for helping build the silicone torso and arm model, Eliot Xing, Kavya Puthuveetil, Akhil Padmanabha, Xingyu Lin, and Daniel Seita for helping provide feedback on the paper. 

\vspace{-0.1cm}
\scriptsize
\bibliographystyle{./IEEEtran}
\bibliography{reference}

\end{document}